\documentclass[10pt,twocolumn,letterpaper]{article}

\usepackage{iccv}
\usepackage{times}
\usepackage{epsfig}
\usepackage{graphicx}
\usepackage{amsmath}
\usepackage{amssymb}
\usepackage{eso-pic}

\usepackage{booktabs}
\newcommand{\PAR}[1]{\vskip4pt \noindent{\bf #1~}}

\usepackage[inline, shortlabels]{enumitem}
\usepackage{soul}
\usepackage{array,calc}
\usepackage{placeins}
\usepackage{multirow}
\usepackage{wrapfig}
\usepackage{color}
\usepackage{colortbl}
\usepackage[dvipsnames]{xcolor}
\usepackage{floatrow}
\usepackage{gensymb}
\usepackage{makecell}
\usepackage{arydshln}

\usepackage[pagebackref=true,breaklinks=true,letterpaper=true,colorlinks]{hyperref}
\usepackage[capitalize]{cleveref}
\crefname{section}{Sec.}{Secs.}
\Crefname{section}{Section}{Sections}
\Crefname{table}{Table}{Tables}
\crefname{table}{Tab.}{Tabs.}

\iccvfinalcopy % *** Uncomment this line for the final submission
\def\iccvPaperID{****} % *** Enter the ICCV Paper ID here

\ificcvfinal\fi

\begin{document}

%%%%%%%%% TITLE
\title{Cross-Dimensional Refined Learning for Real-Time 3D Visual Perception from Monocular Video}

\author{Ziyang Hong \qquad C. Patrick Yue \\
Hong Kong University of Science and Technology \\
% Hong Kong \\
{\tt\small frederick.hong@connect.ust.hk \qquad eepatrick@ust.hk}}

\vspace{-3.2cm}
\twocolumn[{%
\renewcommand\twocolumn[1][]{#1}%
\maketitle
\includegraphics[width=1\linewidth]{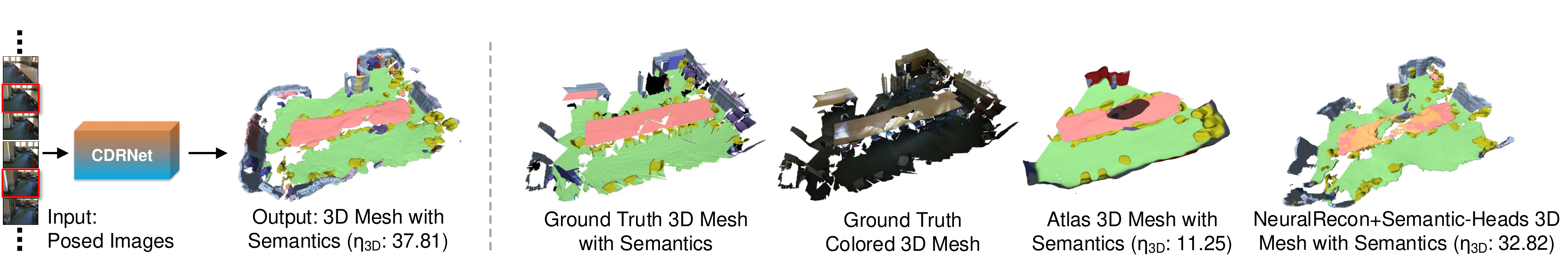}
    \captionof{figure}{
        \textbf{Comparison between the proposed approach and baselines.} 
        Our model is more accurate and coherent in real time, compared to two baseline methods with input from monocular video, Atlas~\cite{murez2020atlas} and NeuralRecon~\cite{sun2021neuralrecon} + Semantic-Heads. Real-time 3D perception efficiency $\eta_{3D}$ the higher the better. Color denotes different semantic segmentation labeling.
        \vspace{15pt}
    }\label{fig:teaser}
}]
\thispagestyle{empty}

% Remove page # from the first page of camera-ready.
% \ificcvfinal\thispagestyle{empty}\fi

%%%%%%%%% ABSTRACT
\begin{abstract}
We present a novel real-time capable learning method that jointly perceives a 3D scene’s geometry structure and semantic labels.
Recent approaches to real-time 3D scene reconstruction mostly adopt a  volumetric scheme, where a Truncated Signed Distance Function (TSDF) is directly regressed.
However, these volumetric approaches tend to focus on the global coherence of their reconstructions, which leads to a lack of local geometric detail.
To overcome this issue, we propose to leverage the latent geometric prior knowledge in 2D image features 
by explicit depth prediction and anchored feature generation, to refine the occupancy learning in TSDF volume.
Besides, we find that this cross-dimensional feature refinement methodology can also be adopted for the semantic segmentation task by utilizing semantic priors.
Hence, we proposed an end-to-end cross-dimensional refinement neural network (CDRNet) to extract both 3D mesh and 3D semantic labeling in real time.
The experiment results show that this method achieves a state-of-the-art 3D perception efficiency on multiple datasets, which indicates the great potential of our method for industrial applications.
\end{abstract}

%%%%%%%%% BODY TEXT
\section{Introduction}
\label{sec:intro}
Recovering 3D geometry and semantics of objects or environment scenes prevails these days with the advent of ubiquitous digitization. 
The digitization of the world where people live can not only help them better understand their environment scenes, but also enable robots to comprehend what they need to know about the world and therewith conducting assigned tasks.
Generally, with surrounding environment measurements as input, 3D reconstruction and 3D semantic segmentation are two key 3D perception techniques~\cite{dahnert2021panoptic, sun2020scalability, han2020live} in the computer vision society, which enable a wide range of applications, including digital twins~\cite{jiang2022ditto, bozic2021neural}, virtual/augmented reality (VR/AR)~\cite{newcombe2011kinectfusion, sun2021neuralrecon}, building information modeling~\cite{mahmud2020boundary, vanegas2010building}, and autonomous driving~\cite{cao2022monoscene, li2022reconstruct}.

Tremendous research efforts have been made for 3D perception techniques. Based on the sensor types, researches on 3D perception can be divided into two main streams, namely active range sensors that capture surface geometry information and RGB cameras that capture texture with perspective projection. 
Originated from KinectFusion~\cite{newcombe2011kinectfusion}, the commodity RGB-D range sensor is used to measure depth data first and then fuse it into Truncated Signed Distance Function (TSDF) volume for 3D reconstruction. Although the follow-up depth-based TSDF fusion methods~\cite{weder2020routedfusion, weder2021neuralfusion, azinovic2022neural, xu2022hrbf, sommer2022gradient} achieve detailed dense reconstruction result, they suffer from global incoherence due to the lack of sequential correlation, the tendency of noise disturbance due to redundant overlapped calculations, and the incapability of semantic deduction due to the lack of texture features.

On the other hand, as camera-equipped smartphones become readily available with built-in inertial measurement units, recent advances have emerged to explore 3D perception with RGB cameras on mobile devices. The problem of reconstructing 3D geometry with posed RGB images input only is referred to as multi-view stereo (MVS).
Existing methods for MVS that are based on deep learning, tend to adopt a volumetric scheme by directly regressing the TSDF volume~\cite{murez2020atlas, stier2021vortx, choe2021volumefusion, sun2021neuralrecon} either as a whole or in fragments. 
However, these volumetric learning methods extract 3D geometric feature representation simply from the back projection of 2D image features, resulting in the mismatch to the 2D information priors for the predicted 3D reconstruction. Moreover, the intrinsic end-to-end learning manner and the lack of local details on the reconstructed mesh of these volumetric schemes result in an inferior semantic deduction based on its 3D reconstruction prediction.

What's worse, these learning-based methods tend to store their entire computational graphs in memory for aggregation and require prohibitive 3D convolution operations~\cite{murez2020atlas, rich20213dvnet, stier2021vortx}, which keeps them from being deployed on robots due to the real-time and low-latency requirements in SLAM. 
% It is necessary to maintain low computation overhead for 3D perception to achieve real-time capability.
These limitations motivate our key idea to utilize 2D explicit predictions to further impose a light-weight feature refinement on the 3D features input in a sparse manner, while keeping the global coherence within the fragments. Unlike these preceding learning-based volumetric works, we conjecture that the utilization of 2D prior knowledge coming out of explicit predictions as a latent feature refinement plays a significant role in learning the feature representation 
in 3D perception. In addition, the feature refinement brought by 2D explicit prediction can be operated within the fragment input for keeping the computation redundancy and thus overhead low, while having the global coherence by correlating different fragments to extract the target 3D semantic mesh.

In this paper, we propose a novel framework, \textit{CDRNet}, to accomplish both 3D meshing and 3D semantic labeling tasks in real-time.
Our key contributions are as follows.
\begin{itemize}[itemsep=0pt,topsep=3pt,leftmargin=*]
    \item We propose a novel, end-to-end trainable network architecture, which cross-dimensionally refines the 3D features with the prior knowledge extracted from the explicit estimations of depths and 2D semantics.
    \item The proposed cross-dimensional refinements yield more accurate and robust 3D reconstruction and semantic segmentation results. We highlight that the explicit estimations of both depths and 2D semantics serve as efficient yet effective prior knowledge for 3D perception learning.
    \item To achieve real-time 3D perception capability, our approach performs both geometric and semantic localized updates to the global map.
    We present a progressive 3D perception system that is capable of real-time interaction with input data streaming from cellphones with a monocular camera.
\end{itemize}

%------------------------------------------------------------------------
\section{Related Work}
\label{sec:related_work}
\begin{figure*}[ht]
    \vspace{-1.2cm}
    \centering
       \includegraphics[width=\linewidth]{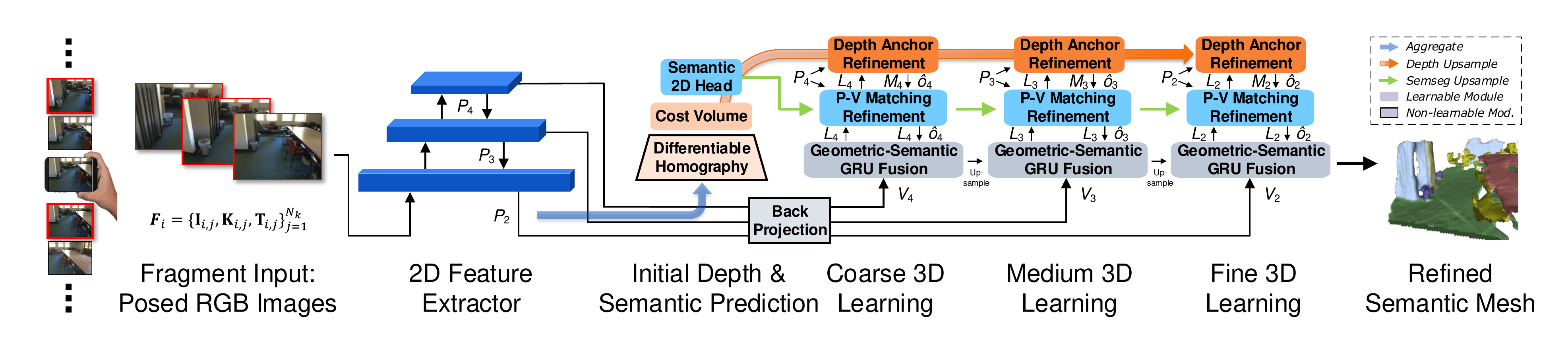}
       \caption{
           \textbf{Overview of CDRNet.} Posed RGB images from monocular videos are wrapped as fragment input for 2D feature extraction, which is used for both depth and 2D semantic predictions for cross-dimensional refinement purposes. 
           To learn the foundational 3D geometry before conducting refinements, the extracted 2D features are back-projected into raw 3D features $\mathcal{V}_s$ in different resolutions without any 2D priors involved.
           At each resolution, after being processed by the GRU, the output feature $L_s$ in the local volume is further fed into {\color{YellowOrange}Depth} and {\color{ProcessBlue}Semantics} refinement modules sequentially to have a 2D-prior-refined feature with better representations.
        %   At the fine stage, the output $\mathbf{S}_t^l$ is used to replace corresponding voxels in the global TSDF volume $\mathbf{S}_t^{g}$, yielding the final reconstruction at time $t$.
           }\label{fig:arch}
      \vspace{-0.2cm}
\end{figure*}
\PAR{Real-Time 3D Perception.} 
The prosperity of deep learning hardwares enables both inference and training at the edge~\cite{lecun20191, hong2022efficient}, thus it consolidates the foundation to deploy more and more learning-based 3D perception techniques in real time. KinectFusion~\cite{newcombe2011kinectfusion} first brought in the concept of handling 3D reconstruction tasks in real time with commodity RGB-D sensors. 
Han et al.~\cite{han2020live} presented a real-time 3D meshing and semantic labeling system similar to our work, however, depth measurements from RGB-D sensors are required as input in their work.
Pham et al.~\cite{pham2019real} built up 3D meshes with voxel hashing, and then fuse the initial semantic labeling with super-voxel clustering and a high-order conditional random field (CRF) to improve labeling coherence.
Menini et al.~\cite{menini2021real} extended RoutedFusion~\cite{weder2020routedfusion} by merging semantic estimation in its TSDF extraction scheme for each incoming depth-semantics pair.
NeuralRecon~\cite{sun2021neuralrecon} adopted sparse 3D convolutions and the gated recurrent unit (GRU) to achieve a real-time 3D reconstruction on cellphones, without the capability of semantic deduction. For depth estimation and semantic segmentation, there are also works achieving real-time processing capability~\cite{wang2018mvdepthnet, narita2019panopticfusion, pham2019real}.

\PAR{Voxelized 3D Semantic Segmentation.} 
The learning of semantic segmentation on the voxelized map started from~\cite{cavallari2016semanticfusion}, which extends TSDF fusion pipeline~\cite{newcombe2011kinectfusion} with per-pixel labels. 3DMV~\cite{dai20183dmv} and MVPNet~\cite{jaritz2019multi} further combined both depth and RGB modalities to train an end-to-end network with 3D semantics for voxels and point clouds, respectively. PanopticFusion~\cite{narita2019panopticfusion} performed map regularization based on adopting a CRF on the predicted panoptic labels. Atlas~\cite{murez2020atlas} utilized its extracted 3D features and passed them to a set of semantic heads for voxel labeling, the pyramid features are proven to have strong semantics at all scales than the gradient pyramid in nature, as proven in~\cite{lin2017feature}. BPNet~\cite{hu2021bidirectional} proposed to have a joint-2D-3D reasoning in an end-to-end learning manner. 
Two derivative works~\cite{menini2021real, huang2021real} of RoutedFusion incorporated semantic priors into their depth fusion scheme 
and removed their routing module for less overhead. 
However, none of these works utilize the prior knowledge within the estimated 2D semantics as a 3D feature refinement.

\PAR{Volumetric 3D Surface Reconstruction.} 
Volumetric TSDF fusion became prevalent for 3D surface reconstruction starting from the seminal work KinectFusion~\cite{newcombe2011kinectfusion} due to its high accuracy and low latency. A follow-up work, PSDF-Fusion~\cite{park2019deepsdf} augmented TSDF with a random variable to improve its robustness to sensor noise. Starting from DeepSDF~\cite{park2019deepsdf}, the learned representations of TSDF using depth input dominates the current fad. These learning-based substitutes~\cite{weder2020routedfusion, weder2021neuralfusion, bozic2021neural, azinovic2022neural, xu2022hrbf, sommer2022gradient, yu2022monosdf} to TSDF fusion achieve impressive 3D reconstruction quality compared to the baseline method with the availability of RGB-D range sensors.

Given the fact that range sensors have relatively higher cost and energy consumption than RGB cameras, MonoFusion~\cite{pradeep2013monofusion} is one of the first works to learn TSDF volume from RGB images by fusing the estimated depth into an implicit model. Atlas~\cite{murez2020atlas} started the trend of learning-based methods by a direct regression on TSDF volume. NeuralRecon~\cite{sun2021neuralrecon} achieved a real-time 3D reconstruction learning capability by utilizing sparse 3D convolutions and recurrent networks with key frames as input. TransformerFusion~\cite{bozic2021transformerfusion} and VoRTX~\cite{stier2021vortx} introduced transformers~\cite{vaswani2017attention} to improve the performance by more relevant inter-frame correlation. 
These learning-based methods prevail thanks to the availability of these general 2D feature extractors, such as FPN~\cite{lin2017feature} and U-Net~\cite{ronneberger2015u}. 2D information in RGB images can be effectively extracted and further utilized for constructing their 3D perception counterparts.

However, the learning of the explicit representations of 2D latent geometric features, such as depths and semantics, is \textbf{typically ignored} by all the prior arts. They only treat the 2D feature as an intermediate in the network and then conduct ray back-projection upon it, without considering the explicit representations for their 3D embodiment, which we found are significant prior knowledge for 3D perception.
To extract depth as the explicit 2D representation, VolumeFusion~\cite{choe2021volumefusion} and SimpleRecon~\cite{sayed2022simplerecon} performed local MVS and further fused it into TSDF volume with its customized network, while 3DVNet~\cite{rich20213dvnet} performed sparse 3D convolutions on the feature-back-projected point cloud. Different from above, our method extracts the 2D representations from light-weight network modules, including a portion of MVSNet~\cite{yao2018mvsnet} for depth and a simple 2D MLP head for 2D semantics, 
to conduct the 3D feature refinements. 
% Similar to~\cite{lengyel2021zero, lin2022deep}, 
The refinement incorporates the geometric and semantic prior information to improve the generalizability of our network by correlating the 2D representations in their 3D counterparts.

To the best of our knowledge, we present the very first learning-based method which uses posed RGB images input only to conduct 3D perception tasks in real time, including 3D meshing and semantic labeling.

%------------------------------------------------------------------------
\section{Methods}
\label{sec:methods}
Given a posed image sequence $\mathbf{I}$, our goal is to extract a 3D mesh model that can represent both \textbf{3D geometry} and \textbf{3D semantic labeling}, i.e., 3D meshing with vertices $\mathcal{K}\in\mathbb{R}^3$, surfaces $\mathcal{G}\in\mathbb{N}^3$, and its corresponding 3D semantic labeling $\mathcal{S}\in \mathbb{N}$.
% Each voxel contains a  and a  of our interest.
We achieve this goal by jointly predicting TSDF value $T\in [-1, 1]$ and semantic label $S\in\mathbb{N}$ for each voxel, and then extracting the mesh with the marching cubes \cite{lorensen1987marching}.
Meanwhile, our proposed method aims at establishing a real-time capable deep learning model for these two 3D perception tasks. To quantitatively evaluate the efficiency of conducting these two tasks simultaneously, we define a 3D perception efficiency metric $\eta_{3D}$ by involving frames per second (FPS) in runtime, as shown in Sec. \ref{subsec:datasets_metrics}.

The proposed network architecture is illustrated in Fig. \ref{fig:arch}.
In Sec. \ref{subsec:fragment}, we introduce the joint fragment learning on depth, 2D semantic category, intermediate TSDF, and occupancy using key frames input, for the following cross-dimensional refinements of TSDF and 3D semantics. For each fragment, the geometric features are progressively extracted in a coarse-to-fine hierarchy using binomial inputs GRU to build the learned representations of 3D.
Sec. \ref{subsec:cdr} describes the cross-dimensional refinements for 3D features that refines 3D features with anchored features and semantic pixel-to-vertex correspondences enabled by the depth and 2D semantic predictions, which helps the learning of not only the TSDF value,
% from the latent geometric information in the extracted features
but also the 3D semantic labeling in a sparsified manner. We also present the implementation details including loss design in Sec. \ref{subsec:implementations}. 
Specifications of the network are elaborated in the supplement.

\subsection{Sparse Joint Fragment Learning in a Coarse-to-Fine Manner}\label{subsec:fragment}
Given the inherent nature of great sparsity in the ordinary real-world 3D scene, we utilize sparse 3D convolutions to efficiently extract the 3D feature from each input scene.
However, the memory overhead of processing a 3D scene is still prohibitive, thus we fragment the whole 3D scene and progressively handle each of them,  to further release the memory burden of holding up the huge 3D volume data.
Inspired by~\cite{murez2020atlas, gu2020cascade, sun2021neuralrecon, stier2021vortx, rich20213dvnet}, we adopt a coarse-to-fine learning paradigm for the sparse 3D convolutions to effectively exploit the representation of 3D features in multiple scales.
In each stage of the hierarchy, the raw features in the fragment bounding volume (FBV) is extracted from a GRU by correlating local features and global feature volume.

\PAR{FBV Construction by Image Features.}
Following~\cite{wang2018mvdepthnet, sun2021neuralrecon}, we select a set of key frames as the input sequences out of a  monocular RGB video by querying on each frame's pose, namely the relative translation and optical center rotation with empirical thresholds, $\theta_{key}$ and $t_{key}$. Key frames $\mathbf{I}$, camera intrinsics $\mathbf{K}$, and transform matrices $\mathbf{T}\in SE(3)$ which is an inversion of the camera pose, are all wrapped into a fragment $\mathbf{F}_i=\{\mathbf{I}_{i,j},\mathbf{K}_{i,j},\mathbf{T}_{i,j} \}_{j=1}^{N_k}$ as the input to the network, where $i$, $j$, and $N_k$ denote the fragment index, the key frame index, and the number of key frames in each fragment, respectively.

Once the fragment $\mathbf{F}_i$ is constructed, it is processed by a 2D feature extractor pyramid to extract image features. In the decoder part of the extractor pyramid, three different resolutions of feature maps are extracted sequentially as $\mathcal{P}_s\in\{P_2, P_3, P_4\}$, where the suffix notation of $P$ denotes the scaling ratio level in $\log_2$ similar to~\cite{lin2017feature}. The extracted feature $\mathcal{P}_s$ is then back-projected into a local 3D volume, according to the projection matrix of each frame in $\mathbf{F}_i$. We hereby define FBV as the current local volume $\mathcal{F}_{s,i}=\{T_{s,i}^{x\times y\times z}, S_{s,i}^{x\times y\times z}\}$ that is conditioned on the pyramid layers $\mathcal{P}_s$, where all the 3D voxels that are casted in the view frustums of current $\mathbf{F}_i$ are included.
% Each voxel contains a TSDF value $T\in [-1, 1]$ and a semantic label $S\in\mathbb{N}$ of our interest.

\PAR{Initial Depth and 2D Semantics Learning.}
With the fine feature $P_2$ as input, we build up differentiable homography fronto-parallel planes for the coarse-level depth prediction $\hat{D}_4$. Likewise, 2D semantics prediction $\hat{S}_4^{2D}$ is extracted with a pointwise convolutional decoder as the 2D semantic head using $P_2$. The resolution gap between the input and output feature map provides generalizability. 
The initial depth estimation is retrieved from the features using a light-weight multi-view stereo network via plane sweep~\cite{yao2018mvsnet}. For each source feature map $x$ in $P_2$, we conduct the planar transformation $\mathbf{x}_j \sim \mathbf{H}_j(d) \cdot x$, where ``$\sim$" denotes the projective equality and $\mathbf{H}_j(d)$ is the homography of the $j^{\text{th}}$ key frame at depth $d$. The $j^{\text{th}}$ homography\footnote{For brevity's sake, the transformation from homogeneous coordinates to Euclidean coordinates in the camera projection is omitted here.} in a given fragment input $\mathbf{F}_i$ is defined as:
\begin{equation}
	\mathbf{H}_j(d)
	= d \cdot \mathbf{K}_j \cdot (\mathbf{T}_j \cdot \mathbf{T}_1^{-1}) \cdot \mathbf{K}_1^T
    \enspace.
\end{equation}
To measure the similarity after conducting homography warping, we calculate the variance cost of $\mathbf{x}_j$ and further process it with an encoder-decoder-based cost regularization network. The output logit from the regularization network is treated as the depth probability on each plane and the \textit{soft argmin}~\cite{yao2018mvsnet} is conducted to have initial depth predictions.

\PAR{Geometric and Semantic GRU Fusion.}
Meanwhile, as the 2D features are extracted in different resolutions, they are back-projected from each of the pyramid level in $\mathcal{P}_s$ into raw geometric 3D features $\mathcal{V}_s\in\{V_2, V_3, V_4\}$, which are further sparsified by sparse 3D convolutions.
To improve the global coherence and temporal consistency of the reconstructed 3D mesh, following~\cite{sun2021neuralrecon}, we first correlate the sparse geometric feature $\mathcal{V}_s$ in the current $\mathcal{F}_{s,i}$ using GRU, with the local FBV hidden states $H_{s,i-1}$ whose information coming from all of the previous fragments $\mathcal{F}_{s,i^\prime}, i^\prime<i$ and coordinates are masked to be the same as $\mathcal{V}_s$. Such correlation outputs a temporal-coherent local feature $L_{s,i}$ for each stage $s$, which is used to generate dense occupancy intermediate $o_{s,i}$, and passed to the 2D-to-3D cross-dimensional refinements. The global feature volume for the entire scene $G_{s,i}$ is fused by $G_{s,i-1}$ and $L_{s,i}$ given the coordinates of $\mathcal{V}_s$ as masks, and update $H_{s,i}$.
Unlike~\cite{sun2021neuralrecon}, we reuse the same parameters in GRU to process the back-projected and upsampled 3D semantic features to generalize better for the semantic prediction $\hat{S}$ in the current FBV.
This is because inputting TSDF and semantic features sequentially into GRU enables its selective fusion across modalities, thus the feature extracted from the hidden state incorporates more semantic information, as pointed out in~\cite{roldao20223d}. 

For the sake of learning 3D features consistently between scales, we update $\mathcal{V}_s$ at each stage by fusing with the upsampled $L_{s+1,i}$. Inspired by the \textit{meta data} mechanism proposed in~\cite{sayed2022simplerecon}, we further concatenate sparse features, with sparse TSDF, occupancy and semantics after masking with $o_{s,i}$, as the meta feature $L_{s+1,i}$ to be upsampled.
We found the inclusion of semantic information in the hidden state of GRU helps build up a good starting point for the upcoming feature refinements, which is verified in the ablation.

\subsection{2D-to-3D Cross-Dimensional Refinements}\label{subsec:cdr}
The raw coherent features from GRUs lack detailed geometric descriptions, leading to unsatisfactory meshing and semantic labeling results. 
% To this aim, 
To overcome these issues, we propose to leverage the 2D feature that is latent after incorporating the learning of depth and semantic frame for the refinement purposes. We notice that with
the learning of depth and 2D semantics, the 2D features now reside in the latent space which can generalize to more accurate
3D geometry and semantics via cross-dimensional refinements.

\PAR{2D-to-3D Prior Knowledge.}
Consider a probabilistic prior in the latent space of the output coherent feature coming from GRU, which accounts for the prior knowledge that the pixel information in both depth predictions and 2D semantic predictions should produce high confidence matching with regard to their own 3D representations. The prior conditioned 3D feature for both perception tasks is defined as: 
\begin{equation}
% \nonumber\\[-37pt]
\label{eq:x-prior}
    X_{prior} = f(L_{s,i})= f\bigl(H_{s,i} (\mathcal{V}_s,H_{s,i-1}\mid\mathcal{F}_{s,i})\bigr)
	 \enspace,
\end{equation}
where $f(\cdot)$ is the 2D-to-3D feature refinement process for either 3D meshing or 3D semantic labeling, whose input is $L_{s,i}$ extracted from $\mathcal{V}_s$ and $H_{s,i-1}$ given $\mathcal{F}_{s,i}$. We borrow the notation of $H_{s,i}$ to be a constructor function $H_{s,i}(\cdot)$ indicating GRU. For each voxel in $\mathcal{F}_{s,i}$, both TSDF and semantic labeling predictions can be formulated as:
\begin{equation}
% \nonumber\\[-37pt]
\label{eq:prior-both}
\hat{I}_{s,i} = \epsilon h\bigl(H_{s,i}(\mathcal{V}_s, H_{s,i-1}\mid\mathcal{F}_{s,i})\bigr) + (1-\epsilon)X_{prior} 
% \\\mathcal{L} = & \zseta_2 H_i + (1-\zeta_2)X_{prior-semseg}
	 \enspace,
\end{equation}  % is it \times symbol here? Relate with the supplement derivation, and highlight the derivation is in the supplement, here.
where $\hat{I}_{s,i}\in\mathcal{F}_{s,i}$ is the refined prediction; $\epsilon$ is a random variable for the respective prior, which is jointly learned by the feature refinement modules representing the
2D-to-3D priors and the GRU network trained with maximum likelihood estimation losses; $h(\cdot)$ is the prediction head. The proof of Eq. (\ref{eq:prior-both}) can be found in the supplement.

The key insight is that the voxels back-projected from either depth prediction or semantic label prediction of the input images has strong evidence on its 3D counterparts.
We hereby define anchored voxels $\alpha_i$, as those voxels in $\mathcal{F}_{s,i}$ that are incorporating all the back-projected depth points, given the fact that the 3D reconstruction task is essentially an inverse problem. 
We propose two progressive feature refinement modules to learn the high confidence of the refined features in latent space such that a more accurate $\hat{I}_{s,i}$ can be extracted with the help of 2D-to-3D prior knowledge.

\begin{figure}[t]
    \centering
    \vspace{-0.2cm}
    \includegraphics[width=\linewidth]{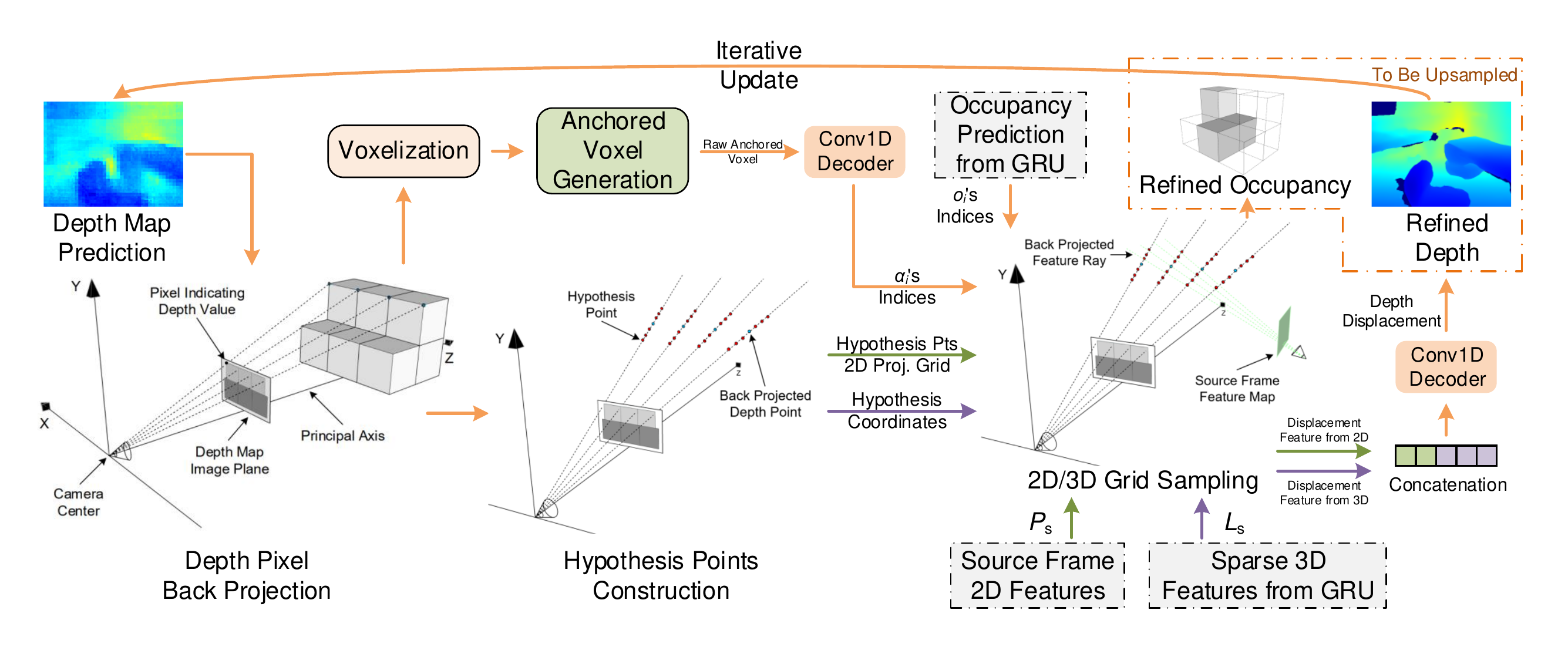}
    \caption{\textbf{Workflow of the depth anchor refinement module.} Anchored voxels are extracted from depth points and further serve as a geometric prior for the occupancy refinement.}
    \vspace{-0.2cm}
    \label{fig:depth_anchor_refmnt}
\end{figure}

\PAR{Depth-Anchored Occupancy Refinement.}
Unlike the volumetric methods~\cite{murez2020atlas, sun2021neuralrecon} that directly regress on the TSDF volume, we propose a novel module in each stage $s$ that can explicitly refine the initial depth, predict depths in resolutions, and further create the 3D anchored features with the depth prediction, as shown in Fig. \ref{fig:depth_anchor_refmnt}. The anchored feature is generated by 3D sparse convolutions with an anchored voxel 
% in the original FBV
on the occupancy intermediate $o_i$
\footnote{The universal stage suffix $s$ is hereinafter omitted for brevity.}.

Intuitively, the anchored voxel has higher confidence of achieving a valid $o_i$ and $T_{s,i}$ close to zero. We imposed the anchored feature on the occupancy feature to reinforce the occupancy information brought by the depth prior.

\begin{figure}[t]
    \centering
    \includegraphics[width=\linewidth]{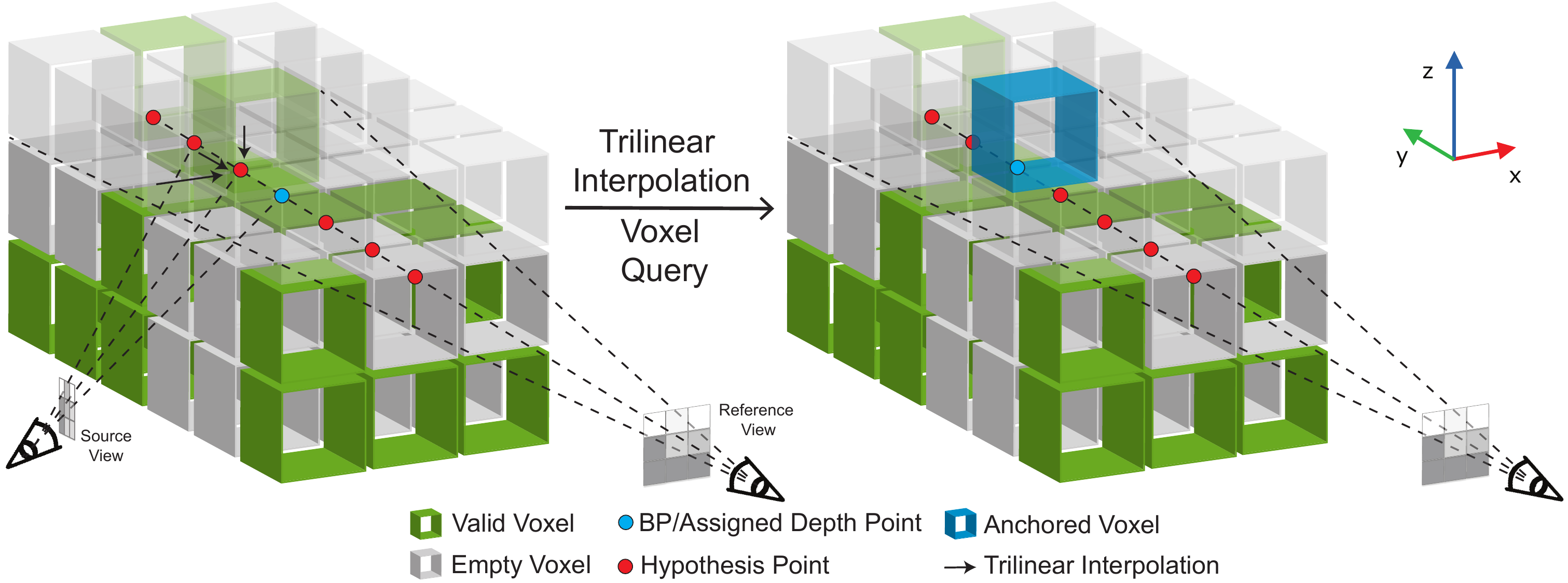}
    \caption{
        \textbf{Anchored voxel generation for occupancy refinement.
        % leveraging the predicted depth based on PointFlow.
        }  % this figure should be anchored feature illustration based on point-flow, what we are seeing here is simply point flow thus misleading people about the over-claim issue...
        An example of occupancy refinement happening on the middle row of a $3\times6\times3$ FBV is shown with geometrically valid voxel highlighted in green. The initial depth prediction is back-projected into FBV and displaced by trilinear interpolation on all depth points, in the range of 6 additional hypothesis points for each depth point. The voxels on the top are set as half transparent for clarity.
        }
        % \color[red]{Description is needed here.}
    \vspace{-0.2cm}
    \label{fig:pointflow}
\end{figure}

Inspired by~\cite{chen2019point, rich20213dvnet}, we conduct PointFlow algorithm for each stage in the coarse-to-fine structure $\mathcal{V}_s$ to determine the depth displacement on the initial depth prediction such that finer depth prediction can be achieved. 
% We define $h$ hypothesis points for each depth pixel. 
Different from the PointFlow algorithm used in~\cite{rich20213dvnet}, we utilize the back-projected depth points from all $N_k$ views in the fragment to query an anchored voxel, which can be further aggregated with $o_i$. Fig. \ref{fig:pointflow} illustrates how these hypothesis points are selected and turned into depth displacement prediction, such that the anchored voxel can be generated.

The anchored voxel index in the 3D volume is sparsified as a mask to update the occupancy prediction as $\hat{o}_i$ in the following:
\begin{equation}
    % \hat{o}^i = o^i \wedge \alpha_i
    \hat{o}_i = o_i \cap \alpha_i
    \enspace.
\end{equation}
The enhanced occupancy prediction $\hat{o}_i$ is used to condition the TSDF volume at the current stage to generate the refined $\hat{T}_i$, which is further sparsified with a light-weight pointwise convolution and upsampled to concatenate with $L_{s,i}$.

\PAR{Pixel-to-Vertex Matching Semantic Refinement.}
In addition to the depth anchor refinement, we propose a semantic cross-dimensional refinement which utilizes the semantic prior that lies in the 2D semantic prediction to have a refined 3D voxel semantic prediction, implemented as follows.
First, the 2D feature backbone learns the 2D semantic prior information that is useful for 3D voxel semantic labeling learning by incorporating the learning of 2D frame semantic labeling.
Second, the sparse 3D feature $L_{s,i}$ is passed to pointwise 3D convolution layers and comes up with the initial 3D voxel semantic labeling predictions in respective scales.
Third, to conduct the semantic feature refinement, we observed that there is a sole 3D voxel counterpart in $\mathcal{F}_{s,i}$ for each pixel on a 2D semantic prediction of $\mathbf{I}_{i,j}$, since 
the surface edges are encoded as vertices.
We define these vertices as the one-on-one matching correspondences to their camera-projected pixels, which are recorded in a matching matrix for masking the 2D features $\mathcal{P}_s$. 

\begin{figure}[t]
    \vspace{-0.2cm}
    \centering
       \includegraphics[width=\linewidth]{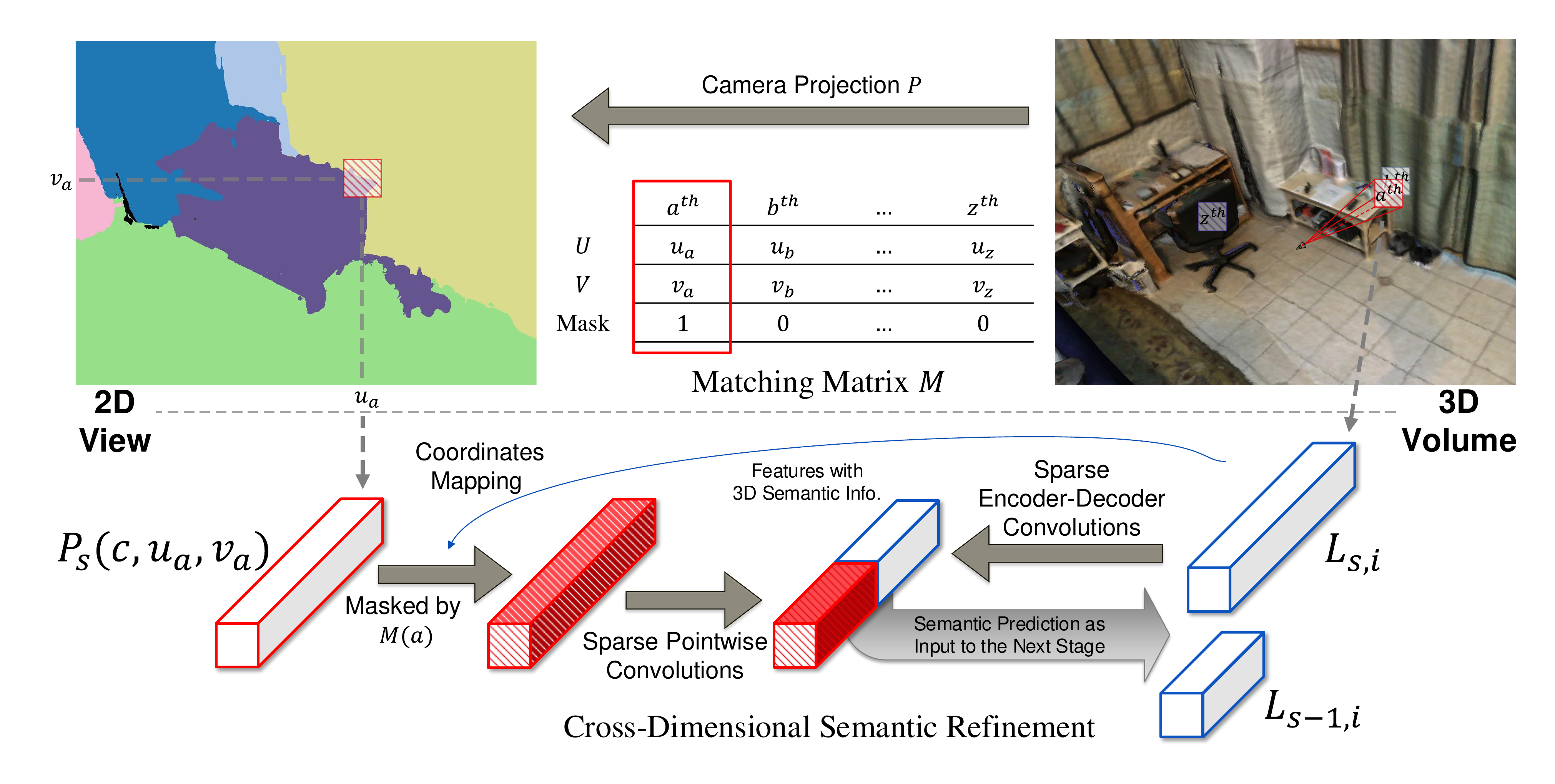}
       \caption{
           \textbf{Workflow of the pixel-to-vertex matching feature refinement.} \textit{Upper:} Matching matrix $\mathbf{M}$ for pixel-to-vertex correspondence is constructed with camera projection. The red-shaded boxes in the 3D volume denote an example of valid correspondence pairs of the 2D semantic prediction $\overrightarrow{m}_a$ and its surrounding 3D scene. The green and purple boxes in the 3D volume view denote the occluded vertex and out-of-view vertex that is not imaged in the 2D semantic prediction, which correspond to $\overrightarrow{m}_b$ and $\overrightarrow{m}_z$, respectively; \textit{Lower:} The 2D features are further masked by $\mathbf{M}(a)$ with the mapped coordinates from the sparse 3D features of the scene that are valid for the current view.
        %   \color{red}{Green, purple and red boxes description. Consistent with the body context.}
        }
       \label{fig:cdr_matching}
      \vspace{-0.3cm}
\end{figure}

The upper part of Fig. \ref{fig:cdr_matching} illustrates the design of the matching matrix that is used to correlate the pixel-vertex pairs for each frame $\mathbf{I}_{i,j}$ across all vertices in $\mathcal{F}_{s,i}$. We construct the matching matrix $\mathbf{M}=\{\overrightarrow{m}_{idx}\}^N_{idx=1}$ for each semantic labeling frame, where $N$ is the number of the vertices in the volume $\mathcal{F}_{s,i}$. Each column of the matching matrix $\mathbf{M}$ is defined as:
\begin{equation}
    \mathbf{M}(idx) = \overrightarrow{m}_{idx} = 
    \begin{bmatrix}
        u_{idx} \\
        v_{idx} \\
        \text{mask}
    \end{bmatrix}.
\end{equation}
For each column, each pixel-vertex pair recorded in the matching matrix, i.e., the $idx^{\text{th}}$ vertex in the 3D volume on the right-hand side of the upper part and its correspondence pixel on the left-hand side is recorded. The last entry of the pixel-vertex pair represents a mask which is recorded as valid when the 2D correspondence for $\mathbf{M}$ is in the current view frustum of the frame.

After the matching matrix $\mathbf{M}$ is constructed, it will be used for masking each of the feature map $\mathcal{P}_s$ with the $\log_2$ scale of $s$ to create a refined feature, whose voxel number is the same as the number of sparse 3D features, 
as shown in the lower part of Fig. \ref{fig:cdr_matching}.
Meanwhile, the coordinates of the sparse 3D features $L_{s,i}$ are mapped as the coordinate of the refined feature. 
By doing so, the underlying semantic information from the $\mathcal{P}_s$ can be incorporated by $L_{s,i}$, such that better 3D semantic prediction can be achieved.
Then we use the sparse pointwise convolution to extract its underlined feature from 2D semantics and concatenate it with $L_{s,i}$ to create $L_{s-1,i}$ with semantic information for the refinement in the next finer stage, so as to ensure the 2D semantic priors to have reliable refinement on the sparse coherent features.

\subsection{Implementation Details}\label{subsec:implementations}
Our model is implemented in PyTorch, trained and tested on an NVIDIA RTX3090 graphics card. We empirically set the optimizer as Adam without weight decay~\cite{loshchilov2018decoupled}, with an initial learning rate of 0.001, which goes through 3 halves throughout the training. The first momentum and second momentum are set to 0.9 and 0.999, respectively.
For key frame selection, following~\cite{wang2018mvdepthnet, sun2021neuralrecon}, we set thresholds $\theta_{key}$, $t_{key}$ and fragment input number $N_k$ as 15 degrees, 0.1 meters, and 9, respectively.
A fraction of FPN~\cite{lin2017feature} is adopted as the 2D backbone with its classifier as MNasNet~\cite{tan2019mnasnet}.
MinkowskiEngine~\cite{choy20194d} is utilized as the sparse 3D tensor library. More details are introduced in the supplement.
\PAR{Loss Design.}
Our model is trained in an end-to-end fashion except for the pre-trained 2D backbone. Since our target is to learn the 3D geometry and semantic segmentation of the surrounding scene given the posed images input, we regress the TSDF value with the mean absolute error (MAE) loss, classify the occupancy value with the binary cross-entropy (BCE) loss and the semantic labeling with cross-entropy (CE) loss as:
% \vspace{-0.2cm}
\begin{align}
% \nonumber\\[-37pt]
\label{eq:loss-3D}
\mathcal{L}_{3D} = &\sum^{4}_{s=2}
	\alpha_s \mathcal{L}_\text{MAE}(T_s, \hat{T}_s) + 
	\lambda \alpha_s \mathcal{L}_\text{BCE}(O_s, \hat{O}_s)\nonumber \\ & + 
	\beta_s \mathcal{L}_\text{CE}(S_s, \hat{S}_s)
	\enspace,
\end{align}
% \vspace{-0.1cm}
where $T$, $S$, and $O$ denote TSDF value, semantic labeling, and occupancy predictions. $\alpha_s$, $\beta_s$, and $\lambda$ are the weighting coefficients in different stages for TSDF volume, semantic volume, and positive weight for BCE loss, respectively. By doing so, the learning process stays most sensitive and relevant to the supervisory signals in the coarse stage, and less fluctuating as the 3D features become finer with the upsampling, after log-transforming the predicted and ground-truth TSDF value following~\cite{murez2020atlas}.

To conduct cross-dimensional refinements, we regress the depth estimation with MAE loss and classify the 2D semantic segmentation with CE loss:
\begin{align}
\label{eq:loss-2D-total}
    \mathcal{L}_{2D} = & \mathcal{L}_\text{MAE}(d_{init}, \hat{D}_{init}) + 
    \mathcal{L}_\text{CE}(S^{2D}_2, \hat{S}^{2D}_2)\nonumber \\ & + 
    \sum^{4}_{s=2} \gamma_s \mathcal{L}_\text{MAE}(D_s, \hat{D}_s)
	 \enspace,
 \end{align}
where $D$ and $\gamma_s$ denote depth and the weighting coefficient for depth estimation in different stages. We further wrap the losses into an overall loss $\mathcal{L}= \mathcal{L}_{3D} + \mu \mathcal{L}_{2D}$, where $\mu$ is the coefficient to balance the joint learning of 2D and 3D.
\begin{figure*}[t]
    \vspace{-1.2cm}
    \centering
	\includegraphics[width=\linewidth]{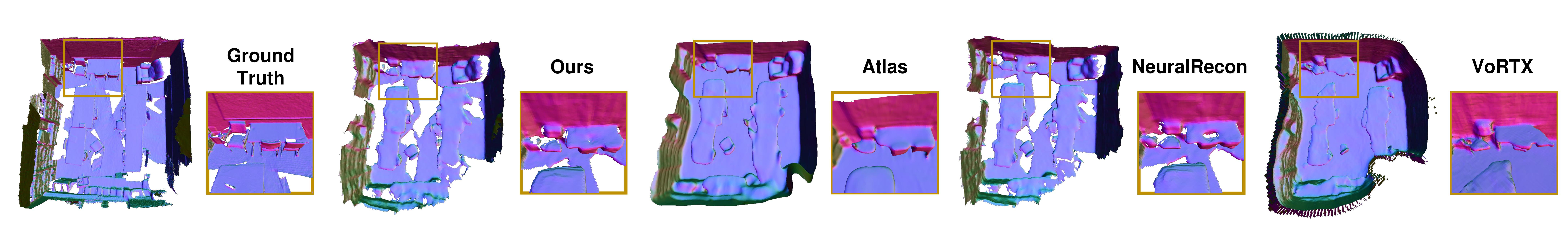}
\caption{\textbf{Qualitative 3D reconstruction results on ScanNet.} Our method is capable of reconstructing consistent and detailed geometry which is neither overly smooth as the one from Atlas~\cite{murez2020atlas} nor eroded with holes as from NeuralRecon~\cite{sun2021neuralrecon}.}
	\label{fig:exp-3dmesh-normal}
    \vspace{-0.2cm}
\end{figure*}

%------------------------------------------------------------------------
\section{Experiments}
\label{sec:experiments}
\subsection{Datasets and Metrics}\label{subsec:datasets_metrics}
\begin{table}[t]
    % \centering
    \vspace{-0.2cm}
    \scriptsize
    \resizebox{1.0\textwidth}{!}{
        \begin{tabular}{ccccccc}
            \Xhline{3\arrayrulewidth}
                        Method                                    & Acc. $\downarrow$  & Comp. $\downarrow$    & Prec. $\uparrow$    & Recall $\uparrow$  & F-Score $\uparrow$  \\           \hline
            Atlas \cite{murez2020atlas}               & 0.124              & 0.074                 & 0.382               & \textbf{0.711}    & 0.499                \\
            NeuralRecon \cite{sun2021neuralrecon}     & 0.073              & 0.106                 & 0.450               & 0.609             & 0.516                \\
            3DVNet \cite{rich20213dvnet}              & \textbf{0.051}     & 0.075                          & \textbf{0.715}               & 0.625             & 0.665                \\
            SimpleRecon \cite{sayed2022simplerecon}   & 0.061              & \textbf{0.055}        & 0.686               & 0.658             & \textbf{0.671}                \\
            VoRTX \cite{stier2021vortx}               & 0.089              & 0.092                  & 0.618 & 0.589 & 0.603\\
            Ours                                      & 0.068              & 0.062                 & 0.609               & 0.616             & 0.612                \\            
    \Xhline{3\arrayrulewidth}
        \end{tabular} }
    \caption{\textbf{Quantitative 3D reconstruction results on ScanNet.} Our method is superior to two main baselines, Atlas and NeuralRecon, and as competitive as other SOTAs on 3D reconstruction.}
    \vspace{-0.2cm}
    \label{tab:scannet-3d-mesh}
\end{table}
We conduct the experiments on two indoor scene datasets, ScanNet (v2)~\cite{dai2017scannet} and SceneNN~\cite{hua2016scenenn}. 
The model is trained on the ScanNet train set, tested and reported on the ScanNet test set and further verified on SceneNN data set.
To quantify the 3D reconstruction and 3D semantic segmentation capability of our method, we use the standard metrics following~\cite{murez2020atlas, sun2021neuralrecon}. Completeness Distance (Comp.), Accuracy Distance (Acc.), Precision, Recall, and F-score, are used for 3D reconstruction, while mean Intersection over Union (mIoU) is used for 3D semantic segmentation.

To evaluate how much robustness a model can achieve while targeting 3D perception tasks in real time, we define the 3D perception efficiency metric $\eta_{3D}=\text{FPS}\times\text{mIoU}\times\text{F-score}$,
since F-score is regarded as the most suitable 3D metric for evaluating 3D reconstruction quality by considering Precision and Recall at the same time~\cite{murez2020atlas,sun2021neuralrecon,sayed2022simplerecon}. It is noteworthy that for fairness across methods, FPS for processing speed is measured in the inference across all captured frames in a given video sequence rather than key frames only, since the input is the same for different methods regardless of their key frame selection scheme.
\begin{table}[t]
    % \centering
    \vspace{-0.2cm}
    \resizebox{1.0\textwidth}{!}{
        \begin{tabular}{ccccccc} 
      \Xhline{3\arrayrulewidth}
      Method        & FPS $\uparrow$   & KFPS $\uparrow$  & FLOPF $\downarrow$ & mIoU $\uparrow$   & $\eta_{3D}$ $\uparrow$       \\ 
      \hline
      3DMV \cite{dai20183dmv}                 & 7.04    & N/A & 65.06G              & 44.2 & N/A                           \\
      BPNet \cite{hu2021bidirectional}        & 4.46    & N/A & 141.06G             & 74.9 & N/A                           \\
      \hline
      Atlas \cite{murez2020atlas}             & 66.3  & N/A                   & 267.04G    & 34.0 & 11.25                              \\
      NeuralRecon \cite{sun2021neuralrecon} + Semantics-Heads  & \textbf{228}                  & \textbf{30.9} & \textbf{42.38G}  & 27.9 & 32.82           \\
      VoRTX \cite{stier2021vortx} + Semantic-Heads              & 119   & 13.5                  & 150.23G    & 13.2  & 9.47                                \\
      Ours                                    & 158   & 21.4                  & 90.62G     & \textbf{39.1}  & \textbf{37.81}   \\ 
      \Xhline{3\arrayrulewidth}
      \end{tabular}
        }
    \caption{\textbf{Quantitative 3D voxel semantic segmentation and overall 3D perception results on ScanNet.} \textit{Upper:} Two representative state-of-the-art methods for semantic segmentation whose input requires either depth or 3D mesh, respectively. No key-frame selection and F-score are involved due to their input modality; \textit{Lower:} RGB-input-only volumetric methods.
    Key-frame FPS (KFPS) is measured with the same selection scheme across all methods. FLOPF is measured with PyTorch operation counter across operations of neural network's learnable modules.
}
    \label{tab:scannet-2d-3d-semantics}
    \vspace{-0.2cm}
\end{table}
\begin{figure*}[ht!]
    \vspace{-1.2cm}
    \centering
	\includegraphics[width=\linewidth]{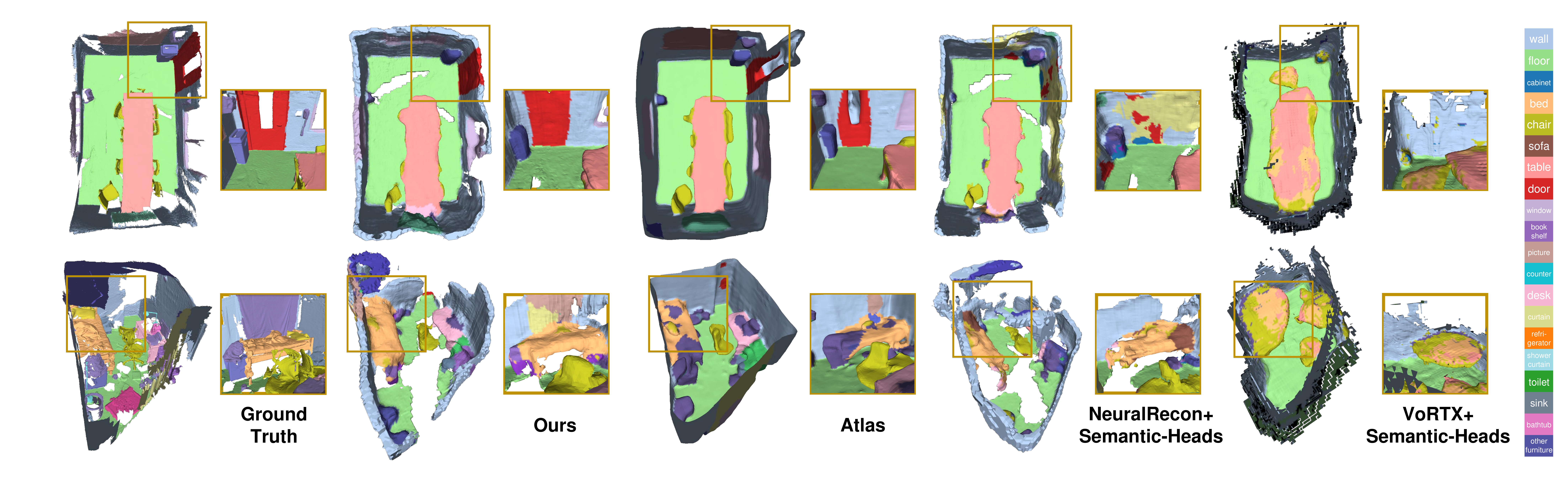}
	\caption{\textbf{Qualitative 3D semantic segmentation results on ScanNet.} Our method consistently outperforms baseline models and sometimes even surpasses the ground-truth labeling, e.g., in the bottom row, the photo-printed curtain above the bed is correctly recognized as ``curtain'' and ``picture'', whereas the ground truth mistakes it as ``other furniture''.}
	\label{fig:exp-3dsemantics}
    \vspace{-0.3cm}
\end{figure*}
\subsection{Evaluation Results and Discussion}\label{subsec:eval_results}
\PAR{3D Perception.}
To evaluate the 3D perception capability, we mainly compare our methods against state-of-the-art works in two categories: volumetric 3D reconstruction and voxelized 3D semantic segmentation methods. 

For 3D reconstruction capability, we compare our proposed method with the canonical volumetric methods~\cite{murez2020atlas, sun2021neuralrecon} and several state-of-the-art 3D reconstruction methods with posed images input~\cite{rich20213dvnet, sayed2022simplerecon}. Fig. \ref{fig:exp-3dmesh-normal} demonstrates the superiority of our method in terms of 3D reconstruction by showing the 3D meshing results in normal mapping. Table \ref{tab:scannet-3d-mesh} shows that our method outperforms two main baseline methods in terms of 3D meshing accuracy.
We further compare both state-of-the-art depth estimation methods and volumetric methods in depth metrics in the supplement to justify from the depth extraction perspective.

\begin{figure}[t]
% 	\centering
% 	\scriptsize
    \vspace{-0.225cm}
	\begingroup
	\setlength{\tabcolsep}{2pt}
	\newcommand{\sz}{0.264}
	\newcommand{\szz}{0.143}
	\hspace{-9pt}
	\resizebox{1.0\textwidth}{!}{
           \includegraphics[width=\linewidth]{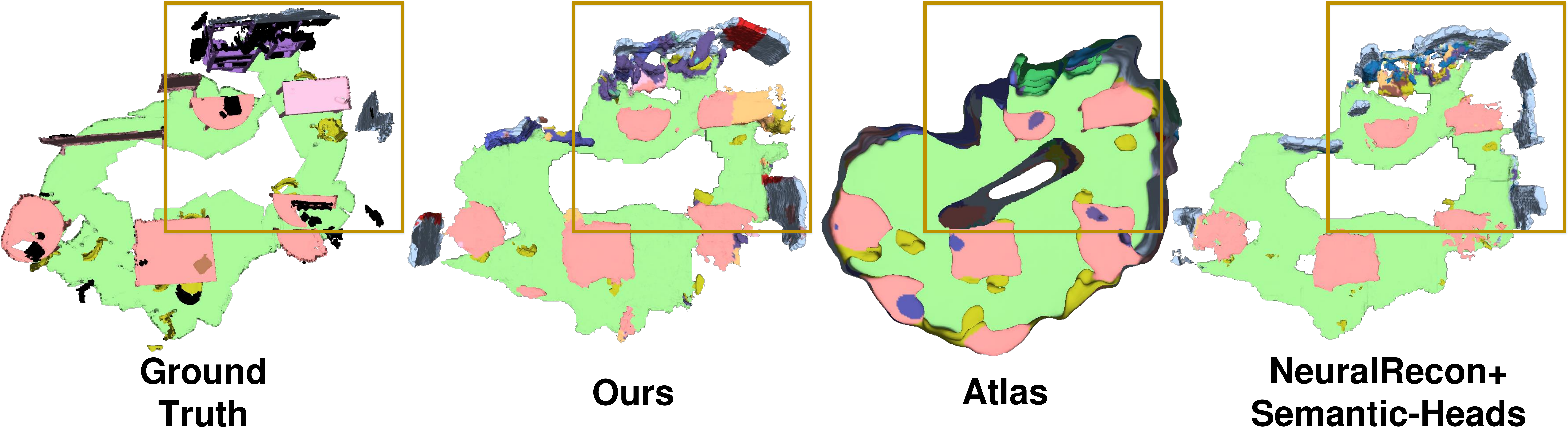}
    }
	\endgroup
    \begingroup	 %
	\resizebox{1.0\textwidth}{!}{
	\begin{tabular}{ccccccc}
        \Xhline{3\arrayrulewidth}
		Method                                          & Acc. $\downarrow$ & Comp. $\downarrow$ & FPS $\uparrow$ & F-Score $\uparrow$ & mIoU $\uparrow$ & $\eta_\text{3D}$ $\uparrow$  \\
        \hline  
		Atlas~\cite{murez2020atlas}                     & 0.074             & 0.164                  & 54.7             & 0.499            & 31.4             & 8.57 \\
        NeuralRecon~\cite{sun2021neuralrecon} + Semantic-Heads     & 0.138             & 0.216                                   & \textbf{178}     & 0.510          & 15.9          & 14.43 \\
		Ours                                            & \textbf{0.068}             & \textbf{0.143}                  & 121            & \textbf{0.611}               & \textbf{36.7} & \textbf{27.09} \\
        \Xhline{3\arrayrulewidth}
	\end{tabular}}
	\endgroup
	\caption{\textbf{Qualitative and quantitative 3D pereception results on SceneNN dataset.} Our method is proven to be generalized to SceneNN without pre-training on the SceneNN train set.}
	\label{fig:exp-3dscenenn}	
    \vspace{-0.2cm}
\end{figure}

For 3D semantic segmentation quality, we compare Atlas, NeuralRecon with semantic heads, and VoRTX with semantic heads with our methods in Table \ref{tab:scannet-2d-3d-semantics}. We augment three stages of MLP heads on top of the flattened 3D features to predict the semantic segmentation for both baselines. Due to its lack of 3D feature extraction, SimpleRecon, as one of the SOTA baselines, is intrinsically incapable of following this modification for semantics as well as being combined with our proposed cross-dimensional refinement techniques.
Table \ref{tab:scannet-2d-3d-semantics} shows that our method outperforms these two baselines. Besides mIoU for semantic segmentation, we include FPS and $\eta_{3D}$ for 3D perception efficiency in the comparison.
We also include two state-of-the-art 3D semantic segmentation methods, 3DMV~\cite{dai20183dmv} and BP-Net~\cite{hu2021bidirectional}. It shows that our method can achieve mIoU results nearly comparable to 3DMV, but with only RGB images as input.
Overall, our method achieves the best 3D semantic segmentation performance and highest 3D perception efficiency among all the volumetric methods.
Fig. \ref{fig:exp-3dsemantics} and Fig. \ref{fig:exp-3dscenenn} illustrate the 3D semantic labeling results. 
We found that the semantic information generation on VoRTX is unsatisfying, 
mostly caused by its bias on geometric features brought by the projective occupancy mentioned in \cite{stier2021vortx}.

\PAR{Efficiency.}
Since our main goal is to achieve real-time processing performance while solving 3D perception tasks, we compare the computational efficiency of our model against other RGB-input-only volumetric methods in Table \ref{tab:scannet-2d-3d-semantics}. The 3D perception efficiency metric $\eta_{3D}$ for several 3D semantic segmentation works are shown there.
We employ FPS, which is commonly used to measure efficiency for
2D-input 3D perception methods~\cite{murez2020atlas,sun2021neuralrecon,stier2021vortx}, as a metric to bring out and emphasize the nature of real-time system.
We also include the floating-point operations per frame (FLOPF) to
compare the learnable parameters' operations across different methods.
The superiority in $\eta_{3D}$ of our method manifests that it has better deployment potential for real-life 3D perception applications. From the human user's and robotic SLAM's points of view, our method greatly surpasses the threshold of being real-time, 90.17 FPS, as elaborated in the supplement. 
It shows that our method is more suitable for real-time industrial scenarios with input data from low-cost portable devices compared to baseline methods.

\subsection{Ablation Study}\label{subsec:ablation_study}
To analyze the effectiveness of cross-dimensional refinement,
we present 3D perception efficiency $\eta_{3D}$ and its components of with different modifications in Table \ref{tab:ablations}. In other experiments above, we adopt (e) as our method.

\noindent\textbf{Binomial GRU Fusion.}
In (a), we remove the back-projected semantics input to GRU in the pipeline. Compared with (e), both F-score and mIoU of the removal degrade since no hidden semantic information from last FBV is fused with GRU anymore. Although FPS increases due to fewer computations, the efficiency $\eta_{3D}$ is worse.

\noindent\textbf{Depth Refinement.}
In (c), we remove the depth anchor refinement in the pipeline. The loss in F-score and mIoU manifests that the geometric feature without depth anchor refinement becomes inferior, which means depth anchor refinement can improve 3D reconstruction performance.
\begin{table}[t]
    % \centering
    \vspace{-0.2cm}
    \resizebox{1.0\textwidth}{!}{
        \begin{tabular}{cccccccccc}
            \Xhline{3\arrayrulewidth}
           & \multirow{2}{*}{GRU Input} & \multicolumn{2}{c}{Depth} & \multicolumn{2}{c}{Semantics} & \multirow{2}{*}{F-Score$\uparrow$} & \multirow{2}{*}{mIoU$\uparrow$}      & \multirow{2}{*}{FPS $\uparrow$}     & \multirow{2}{*}{$\eta_{3D}\uparrow$} \\ \cline{3-6}
         &                  & DE & AR           & SE & PVR \\ \hline
          (a)  & Geo.         & \checkmark & \checkmark & \checkmark & \checkmark  & 0.477              & 31.7                 & 190                & 28.73                       \\
          (b)  & Geo.+ Sem.   & \checkmark &            & \checkmark &             & 0.479              & 27.1                 & \textbf{232}               & 30.12                            \\
          (c)  & Geo.+ Sem.   & \checkmark &            & \checkmark & \checkmark  & 0.482              & 34.5                 & 169                & 28.10                       \\
          (d)  & Geo.+ Sem.   & \checkmark & \checkmark & \checkmark &             & 0.556              & 26.8                 & 226       & 33.68                       \\
          (e)  & Geo.+ Sem.   &\checkmark  & \checkmark & \checkmark & \checkmark  & \textbf{0.612}     & \textbf{39.1}        & 158                & \textbf{37.81}               \\
            \Xhline{3\arrayrulewidth}
        \end{tabular} }
    \caption{\textbf{Ablation study.} We assess our method by removing each of the proposed feature fusion techniques on ScanNet. DE, AR, SE, and PVR denote depth estimation, anchored refinement, 2D semantics estimation, and point-to-vertex refinement, respectively.}
    \vspace{-0.2cm}
    \label{tab:ablations}
\end{table}

\noindent\textbf{Semantic Refinement.}
We validate the semantic refinement in the pipeline by removing this module and, as shown in (d). The mIoU drops due to the insufficient learning information from semantic heads only. This result demonstrates the effectiveness of our semantic refinement scheme based on pixel-to-vertex matching for improving 3D semantic segmentation performance. We also experiment with no refinements but depth and 2D semantics learning setup in (b), which gives the highest FPS but not satisfying 3D perception performance.

%------------------------------------------------------------------------
\vspace{-0.15cm}
\section{Conclusion}
\label{sec:conclusion}
In this paper, we proposed a lightweight volumetric method, \textit{CDRNet}, that leverages the 2D latent information about depths and semantics as the feature refinement to handle 3D reconstruction and semantic segmentation tasks effectively. We demonstrated that our method has real-time 3D perception capabilities, and justified the significance of utilizing 2D prior knowledge when solving 3D perception tasks.
% Extensive experiments on various datasets
Experiments on multiple datasets justify the 3D perception performance improvement of our method compared to prior arts. From the application point of view, the scalability of \textit{CDRNet} supports the notion that 2D priors should not be disregarded in 3D perception tasks and opens up new avenues for achieving real-time 3D perception using input data from readily accessible portable devices such as smartphones and tablets.

\PAR{Acknowledgements:}
This work is in part supported by Bright Dream Robotics (BDR) and the HKUST-BDR Joint Research Institute Funding Scheme under Project HBJRI-FTP-005 (OKT22EG06).

{\small
\bibliographystyle{ieee_fullname}
\bibliography{egbib}
}

\end{document}